\RequirePackage{fix-cm}
\documentclass[twocolumn]{svjour3}          
\smartqed  
\usepackage{graphicx}
\usepackage{natbib}
\usepackage{multirow}
\usepackage{amsmath}
\usepackage{amssymb}
\usepackage{bbm}
\usepackage{bm}
\usepackage{booktabs}
\usepackage{array} 
\usepackage{blindtext}
\usepackage{comment}
\usepackage{subcaption}

\usepackage{pifont}
\newcommand{\cmark}{\ding{51}}
\newcommand{\xmark}{\ding{55}}

\usepackage[dvipsnames]{xcolor}
\usepackage{color, colortbl}

\def\eg{\emph{e.g.}} 
\def\ie{\emph{i.e.}}

\definecolor{remark}{rgb}{1,.5,0} 
\definecolor{citecolor}{rgb}{0,0.443,0.737} 
\definecolor{linkcolor}{rgb}{0.956,0.298,0.235} 
\definecolor{gray}{gray}{0.5}
\definecolor{cyan}{rgb}{0.831,0.901,0.945}
\definecolor{teal}{rgb}{0,0.5,0.5}
\definecolor{lightskyblue}{rgb}{0.53,0.8,0.976}

\usepackage{xspace}
\newcommand{\ours}{SHADE\xspace}

\usepackage{amsmath}

\usepackage[pagebackref,breaklinks,colorlinks,bookmarks=false]{hyperref}

\colorlet{dark-blue}{blue!70!black}
\colorlet{dark-green}{green!60!black}
\colorlet{dark-red}{red!80!black}
\definecolor{mypink}{RGB}{219, 48, 122}
\hypersetup{
 colorlinks=true,%
 citecolor=citecolor,%
 filecolor=citecolor,%
 linkcolor=linkcolor,%
 urlcolor=magenta
}

\makeatletter
\renewcommand\paragraph{
  \@startsection{paragraph} 
  {4} 
  {\z@} 
  {.5em \@plus1ex \@minus.2ex} 
  {-.5em} 
  {\normalfont\normalsize\bfseries} 
}
\makeatother
\begin{document}
\sloppy

\title{
Style-Hallucinated Dual Consistency Learning: A Unified Framework for Visual Domain Generalization
}


\author{Yuyang Zhao         \and
        Zhun Zhong \and
        Na Zhao \and
        Nicu Sebe \and
        Gim Hee Lee
}

\institute{Yuyang Zhao \at
              Department of Computer Science, National University of Singapore, Singapore \\
              \email{yuyang.zhao@u.nus.edu}
          \and
          Zhun Zhong (Project Lead) \at
              School of Computer Science, University of Nottingham, United Kingdom \\
              \email{zhunzhong007@gmail.com}
          \and
          Na Zhao \at
          Information Systems Technology and Design Pillar,
             Singapore University of Technology and Design, Singapore \\
              \email{na\_zhao@sutd.edu.sg}
          \and
          Nicu Sebe \at
              Department of Information Engineering and Computer Science, University of Trento, Trento, Italy \\
              \email{sebe@disi.unitn.it}
            \and
          Gim Hee Lee (Corresponding Author) \at
            Department of Computer Science, National University of Singapore, Singapore \\
            \email{gimhee.lee@nus.edu.sg}
}

\date{Received: date / Accepted: date}

\maketitle

\begin{abstract}
Domain shift widely exists in the visual world, while modern deep neural networks commonly suffer from severe performance degradation under domain shift due to the poor generalization ability, which limits the real-world applications. The domain shift mainly lies in the limited source environmental variations and the large distribution gap between source and unseen target data. 
To this end, we propose a unified framework, \textbf{S}tyle-\textbf{HA}llucinated \textbf{D}ual consist\textbf{E}ncy learning (\textbf{\ours}), to handle such domain shift in various visual tasks.
Specifically, \ours is constructed based on two consistency constraints, Style Consistency (SC) and Retrospection Consistency (RC). SC enriches the source situations and encourages the model to learn consistent representation across style-diversified samples. 
RC leverages general visual knowledge to prevent the model from overfitting to source data and thus largely keeps the representation consistent between the source and general visual models. 
Furthermore, we present a novel style hallucination module (SHM) to generate style-diversified samples that are essential to consistency learning. 
SHM selects basis styles from the source distribution, enabling the model to dynamically generate diverse and realistic samples during training. 
Extensive experiments demonstrate that our versatile \ours can significantly enhance the generalization in various visual recognition tasks, including image classification, semantic segmentation and object detection, with different models, \ie, ConvNets and Transformer.
Code is available at \url{https://github.com/HeliosZhao/SHADE-VisualDG}.

\end{abstract}


\section{Introduction}\label{sec:intro}

\begin{figure*}[t]
    \centering
    \includegraphics[width=.95\linewidth]{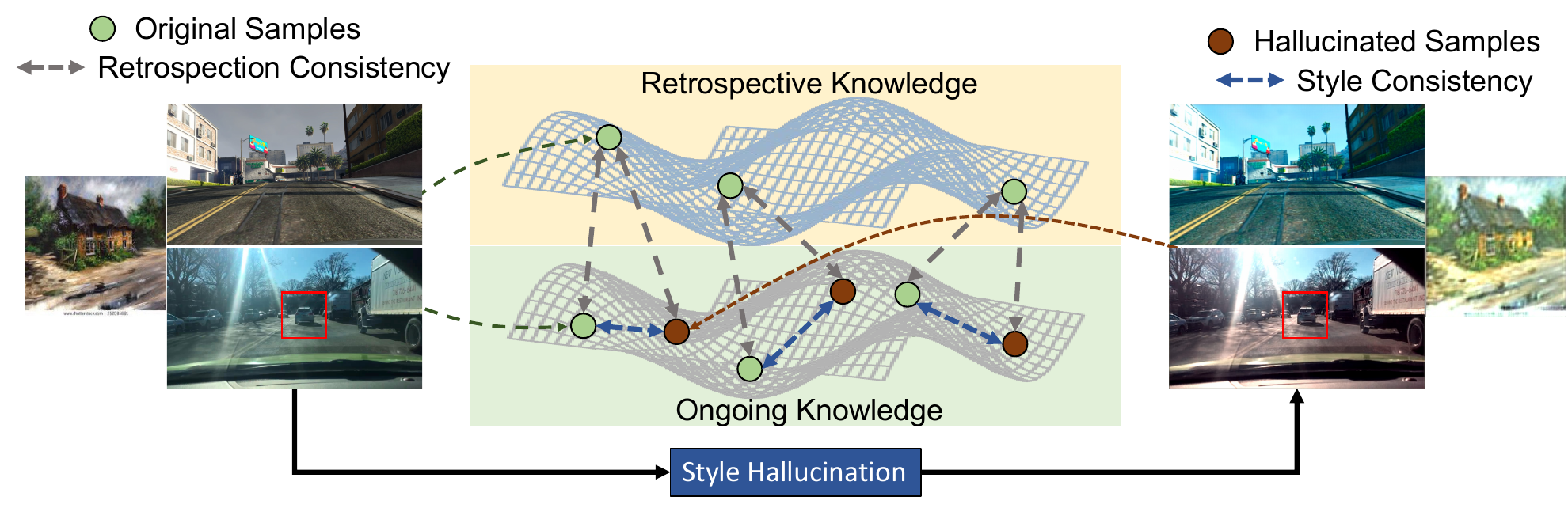}
    \caption{Illustration of the proposed dual consistency constraints for three visual tasks. We generate hallucinated samples (\textcolor{brown}{brown circle}) from the style hallucination module and then utilize the paired samples and general visual (retrospective) knowledge to learn style consistency (\textcolor{dark-blue}{blue dash line}) and retrospection consistency (\textcolor{gray}{gray dash line}).}
    \label{fig:intro}
\end{figure*}

With the development of deep neural networks and the introduction of abundant annotated data, fully-supervised methods have achieved remarkable success in various visual recognition tasks, including but not limited to image classification~\citep{he2016deep,dosovitskiy2020image,liu2021swin}, object detection~\citep{ren2015faster,he2017mask,carion2020end} and semantic segmentation~\citep{hoffman2016fcns,deeplab,xie2021segformer}. These visual recognition tasks are the fundamental and crucial components of the computer vision world. However, such significant achievements  
heavily reply on the availability of large-scale annotated data%
, which are expensive and time-consuming to collect, especially for semantic segmentation and object detection.
For example, it takes more than 1.5 hours to annotate the semantic labels of a 1024$\times$2048 driving scene~\citep{cityscapes}, and the time is even doubled for scenes under adverse weather~\citep{acdc} and poor lighting conditions~\citep{darkzurich}. 
In addition, even though given abundant labeled training data, the significant performance of the deep learning model is limited to independent and identically distributed (\textit{i.i.d}) datasets. Nevertheless, out-of-distribution (OOD) data that are totally unseen during training are inevitably in the real-world applications, \eg, weather change, and the models commonly suffer from catastrophic performance degradation when facing unseen situations.

To alleviate the heavy annotation cost and distribution shift,
domain generalization (DG)~\citep{zhou2021mixstyle,wu2022single,advstyle} is introduced in the community. 
DG only leverages annotated source data to train a robust model that can cope with different unseen conditions. 
The source domain can be the annotated real-world data but can also be the synthetic data from pre-designed engine~\citep{synthia,gtav}, where the later can greatly reduce the annotation cost.
In view of the practicality of DG, previous works have been independently investigating it in image classification~\citep{zhou2021mixstyle}, semantic segmentation~\citep{advstyle} and object detection~\citep{wu2022single}. 
In this paper, we aim to propose a unified and versatile framework that is applicable to the above three visual recognition tasks.

The main challenge for DG is to cope with the significant domain shift between source and unseen target domains, which can be roughly divided into two aspects. First, the environmental variations and diversity in source data are very limited compared to those of unseen target data. 
Second, there exists large distribution gap between the source and target data, \textit{e.g.,} image styles and characteristics of objects.
To learn the domain-invariant model that can address the domain shift, previous works mainly focus on three aspects: 
(1) designing tailor-made modules~\citep{robustnet,ibn,wu2022single} to remove domain-specific information; (2) leveraging extra data to transfer source data~\citep{FSDR,DRPC} to possible target styles for narrowing the distribution gap; (3) diversifying source data within the domain via style augmentation~\citep{zhou2021mixstyle,wang2021L2D} or adversarial perturbation~\citep{advstyle,shankar2018generalizing}.
However, the removal of domain-specific information is not complete and explicit due to the lack of target information; the extra style transfer heavily relies on extra data, which are not always available in practice, and ignores the invariant representation within the source domain.
Taking the above into account, in this paper, we follow the third paradigm to diversify samples in the source domain. In addition, we explicitly introduce two constraints to help the model effectively learn domain-invariant representation and narrow the domain gap.

To this end, we introduce a novel dual consistency learning framework that can jointly address the above two types of domain shift.
As shown in Fig.~\ref{fig:intro}, we introduce two consistency constraints, \textit{style consistency} (SC) and \textit{retrospection consistency} (RC).
SC encourages the model to learn style invariant representation by forcing the consistency between the samples before and after style variation. RC aims at leading the model less overfitting to the source data with the help of general visual knowledge. Specifically, we leverage the ImageNet~\citep{imagenet} pre-trained model which is available acquiescently in all DG models.
The features from the pre-trained model can reflect the representation in the context of general visual world and thus can serve as the guidance for the ongoing model to retrospect what the visual world looks like and to lead the model less overfitting to the source data.

Style diversifying is crucial for the success of dual consistency learning, and we adopt the style features, \ie, channel-wise mean and standard deviation, to generate new data.
Compared with directly transferring the whole image (\eg, CycleGAN~\citep{cyclegan}), changing style features can maintain the pixel alignment to the utmost extent, which is better for pixel-level tasks like semantic segmentation. 
Previous works~\citep{crossnorm,zhou2021mixstyle} commonly mix or swap styles within the source domain, which will generate more samples of the dominant styles (\eg, daytime in GTAV~\citep{gtav}). Nevertheless, it is not the best way since the target styles may be quite different from the dominant styles. 
To fully take advantage of all the source styles, we propose style hallucination module (SHM), which leverages $C$ basis styles to represent the style space of $C$ dimension and thus to generate new styles. 
Ideally, the basis styles should be linearly independent so the linear combination of basis styles can represent all the source styles. 
However, many unrealistic styles that impair the model training are generated when we directly take $C$ orthogonal unit vectors as the basis. To reconcile diversity and realism, we use farthest point sampling~(FPS)~\citep{qi2017pointnet++} to select $C$ styles from all the source styles as basis styles. 
Such basis styles contain many rare styles since rare styles are commonly far away from the dominant ones. With these basis styles that represent the style space in a better way, we utilize linear combination to generate new styles.
To summarize, we propose the \textbf{S}tyle-\textbf{HA}llucinated \textbf{D}ual consist\textbf{E}ncy learning (\textbf{\ours}) framework for domain generalization in visual recognition tasks, and our contributions are as follows:
\begin{itemize}
    \item We propose the dual consistency constraints for visual domain generalization, which can learn the style invariant representation among diversified styles and narrow the domain gap between source and target domains by retrospecting the general knowledge of the visual world.
    \item We propose the style hallucination module to generate new and diverse styles with the representative basis styles, effectively facilitating the dual consistency learning.
    \item The proposed \ours is generic and can be applied to various domain generalized visual recognition tasks, including image classification, semantic segmentation, and object detection. In addition, \ours can be applied to both ConvNets and Transformer to improve generalization ability.
\end{itemize}

An earlier and preliminary study of \ours was published in ECCV 2022~\citep{zhao2022shade}. 
In comparison, this paper further introduces the following significant contributions.
Specifically, 
(1) we investigate the generalization ability of Transformer-based semantic segmentation model and demonstrate that many state-of-the-art DG methods cannot work well on this model but \ours can still achieve significant performance;
(2) we propose the general formulation of dual consistency learning, which enables us to apply it to different visual tasks with a light modification; 
(3) we demonstrate that \ours is complementary to different state-of-the-art DG method in image classification, leading to further improvement; 
(4) we demonstrate that \ours is applicable to domain generalized object detection to improve the robustness under the environmental change in the urban scenes; 
(5) more ablation experiments are provided to further verify the effectiveness of each component of our method;
(6) visualizations on the three visual tasks are provided to better understand the effect of our SHM.

\begin{figure*}[t]
    \centering
    \includegraphics[width=.95\linewidth]{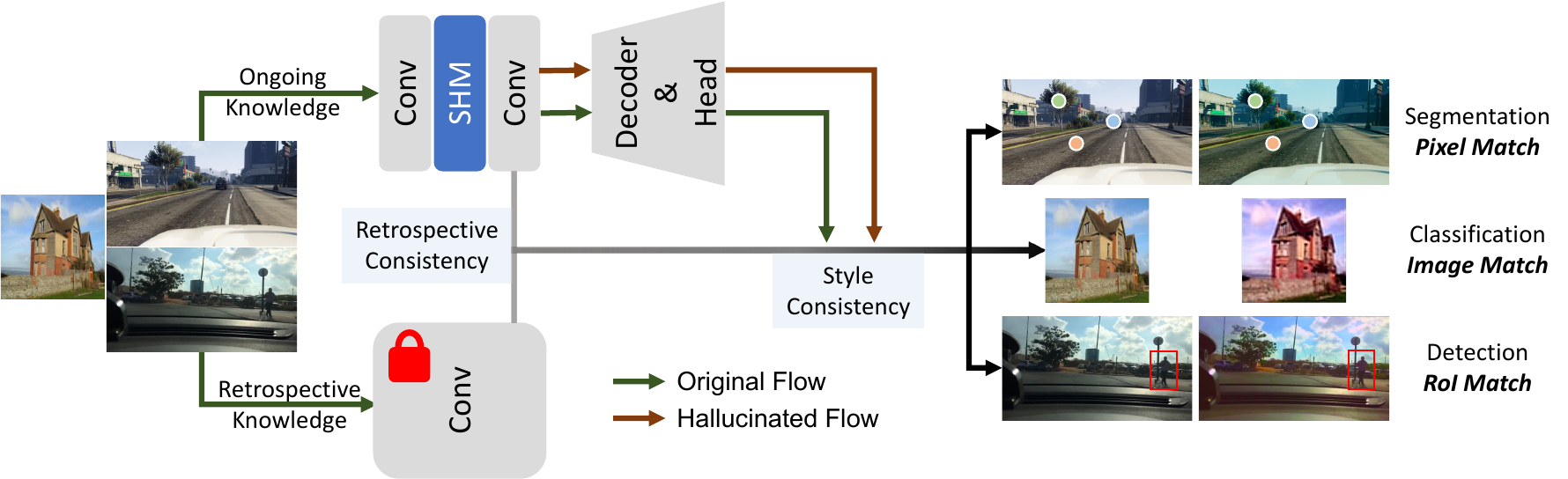}
    \caption{Overview of the proposed style-hallucinated dual consistency learning (\ours) framework. Training images are fed into the training model (ongoing knowledge) and the ImageNet pre-trained model (retrospective knowledge). 
    The style hallucination module is inserted into a certain layer of the training model to generate stylized samples.
    Finally, the model is optimized by the dual consistency losses: style consistency $\mathcal{L}_{\text{SC}}$ and retrospection consistency $\mathcal{L}_{\text{RC}}$. Note that the standard task loss is also used and we omit it here for brevity.}
    \label{fig:framework}
\end{figure*}

\section{Related Work}

\noindent\textbf{Domain Generalization.} 
To tackle the performance degradation in the out-of-distribution conditions, domain generalization (DG)~\citep{FSDR,DRPC,robustnet,zhou2021mixstyle,zhao2021learning,gong2021dlow,yuan2022domain,du2022cross} is introduced to learn a robust model with one or multiple source domains, aiming to perform well on unseen domains. 
The domain generalization methods can be divided into two parts based on the number of source domains, \ie, single-source DG and multi-source DG. In general, domain generalized model requires to learn the domain-invariant representation. Thus, multi-source DG can explicitly learn the domain-invariant representations from multiple source domains~\citep{li2018deep,li2018learning,zhao2021learning} or generating samples across source domains~\citep{zhou2021mixstyle,nuriel2021permuted}.
Single-source DG is more complex, which can only investigate the robust representation from one source domain. The mainstream of single-source DG is to diversify the source sample with additional styles~\citep{wang2021L2D,crossnorm} or perturbation~\citep{qiao2020learning,zhao2020maximum}.

In semantic segmentation, considering the expensive annotation cost, synthetic data are commonly adopted as the source domain in recent domain generalized semantic segmentation (DG-Seg) works. 
To address the domain shift, one mainstream of previous works~\citep{FSDR,DRPC} focuses on diversifying training data with real-world templates and learning the invariant representation from all the domains. Another mainstream aims at directly learning the explicit domain-invariant features within the source domain~\citep{robustnet,ibn,san}. 
IBN-Net~\citep{ibn} and ISW~\citep{robustnet} leverage tailor-made instance normalization block and whitening transformation to reduce the impact of domain-specific information. 
More recently, style augmentation based methods~\citep{advstyle,lee2022wildnet} attract more attention in DG-Seg. 

Domain generalized object detection (DG-Det) is a relatively new setting. The mainstream of DG-Det is disentangling domain-invariant features from the domain-specific features.
\cite{lin2021domain} leverage feature disentangle and representation reconstruction from two source domains to learn the generalizable detection model.
\cite{wu2022single} focus on the single-source DG-Det in the urban scenes, where they disentangle the domain-invariant and domain-specific information via cyclic operation and use self-distillation to improve the generalization ability.

In this paper, we focus on the visual domain generalization in image classification, semantic segmentation and object detection. We mainly focus on the single-source setting but also demonstrate the effectiveness of our method under the multi-source situations.

\noindent\textbf{Consistency learning (CL).}
CL is adopted by many computer vision tasks and settings. One main stream is leveraging CL to exploit the unlabeled samples in semi-supervised learning~\citep{french2019semi,tarvainen2017mean}, unsupervised domain adaptation~\citep{zhou2022uncertainty,zhou2021context,chen2021scale,zheng2021rectifying,zheng2020unsupervised,zheng2022adaptive,chen2022pipa} and novel class discovery~\citep{zhao2022ncdss,fini2021unified,roy2022class,zhong2021openmix}.
CL is also applied to address the corruptions and perturbations~\citep{hendrycks2020augmix,wang2021augmax} by maximizing the similarity between different augmentations.
In addition, CL is also used in self-supervised learning~\citep{simclr,moco} as the contrastive loss to utilize totally unlabeled data.
We introduce CL into domain generalization, leading the model robust to various styles. We also leverage consistency with general visual knowledge to prevent the model from overfitting to the source data. 

\noindent\textbf{Style Variation.}
Style features are widely explored in style transfer~\citep{dumoulin2016learned,adain}, which aims at changing the image style but maintaining the content. 
Inspired by this, recent domain generalization methods leverage the style features to generate diverse data of different styles to improve the generalization ability.
Swapping~\citep{crossnorm,zhao2022sfocda} and mixing~\citep{zhou2021mixstyle} existing styles within the source domains is an effective way and generating new styles~\citep{wang2021L2D} by specially designed modules can also make sense. More recently, AdvStyle~\citep{advstyle} applies adversarial learning to the style features to generate diverse and hard samples.
We also only leverage the styles within the source domain but take the relatively rare styles in the source domain into account, thus generating more diverse samples to improve generalization ability.

\section{Methodology}
\noindent\textbf{Preliminary.} In domain generalization, one or multiple labeled source domains $\mathcal{S}=\{x_S^i, y_S^i\}_{i=i}^{N_S}$, where ${N_S}$ is the number of source domains, are used to train a recognition model, which is deployed to unseen target domains $\mathcal{T}$ directly. 
In general, the source and target domains share the same label space $Y_S, Y_T \in (1, N_C)$ but belong to different distributions. The goal of this task is to improve the generalization ability of model in unseen domains using only the source data.

\noindent\textbf{Overview.} 
To solve the above challenging problem, we propose the \textbf{S}tyle-\textbf{HA}llucinated \textbf{D}ual consist\textbf{E}ncy learning (\textbf{\ours}) framework, which is quipped with dual consistency constraints and a Style Hallucination Module (SHM). The dual consistency constraints effectively learn the domain-invariant representation and ameliorate the overfitting issue via general visual knowledge. SHM enriches the training samples by dynamically generating diverse styles, which catalyzes the advantage of dual consistency learning. The overall framework is shown in Fig.~\ref{fig:framework} and SHM is illustrated in Fig.~\ref{fig:shm}.

\subsection{Dual Consistency Constraints}
In \ours, we introduce two consistency learning constraints: (1) Style Consistency (SC) that aims at learning the consistent predictions across stylized samples. (2) Retrospection Consistency (RC) that focuses on narrowing the distribution shift between source data and general visual knowledge in the feature-level, which can alleviate the model overfitting to the source data.

\noindent\textbf{Style Consistency (SC).}
Cross entropy constraint is one of the most important criterion for recognition tasks, focusing on the high-level semantic information for each class.
Instead of the global representation for each class, logit pairing highlights the most invariant information across paired samples, which has demonstrated its effectiveness in learning adversarial samples~\citep{hendrycks2020augmix,kannan2018adversarial}.
Inspired by this, we propose SC to ameliorate the style shift with logit pairing.
To fulfill this, SC requires the style-diversified samples $\Tilde{x}_S$ which share the same semantic content with the source samples $x_S$ but of different styles. 
To generate stylized samples online and maintain the pixel-level semantic information, non-geometry style variation, \eg, MixStyle~\citep{zhou2021mixstyle}, CrossNorm~\citep{crossnorm} and the proposed style hallucination in Sec.~\ref{sec:sh}, is used to obtain the style-diversified samples $\Tilde{x}_S$.
\textbf{Note} that we focus on three visual recognition tasks, including image classification, semantic segmentation and object detection. As shown in Fig.~\ref{fig:framework}, the implementation of style consistency depends on the task.
In general, we minimize the Jensen-Shannon Divergence (JSD) between the posterior probability
of the semantically aligned $\Tilde{x}_S$ and $x_S$:
\begin{equation}
\small
\begin{aligned}
    \mathcal{L}_{\text{SC}} (x_S, \Tilde{x}_S) &= JSD\left(P; \Tilde{P} \right) \\
    &= \frac{1}{2} \left(D_{\text{KL}} [P || Q] + D_{\text{KL}} [\Tilde{P} || Q] \right), \\
\end{aligned}
\end{equation}
where $Q=( P + \Tilde{P} )/2$ is the average information of the original and style-diversified samples. $D_{\text{KL}}$ denotes the KL Divergence between the posterior probability $\{ P, \Tilde{P} \}$ and $Q$. 
JSD highlights the invariant semantic information across two styles, impelling the model to be stable and insensitive to varied styles.

\noindent\textbf{Retrospection Consistency (RC).}
Backbones of generalizable models commonly start from ImageNet~\citep{imagenet} pre-trained weights since the pre-trained backbones have learned general representation of the visual world.
However, the model learns more task-specific information and fit to the source data after training, which limiting the performance on the unseen scenarios.
Since ImageNet pre-trained weights are available to every generalizable model as the initialization, we propose RC to leverage such knowledge to lead the model less overfitting to the source data and to retrospect the knowledge of the visual world lying in initialization.
Specifically, RC is implemented as the feature-level distance minimization between the training model $\theta_S$ and pre-trained model $\theta_{IN}$.
In addition, the style-diversified samples $\Tilde{x}_S$ in style consistency is also used in RC, which can lead the generated samples close to the real-world style. Similar to style consistency, the backbone feature used for RC depends on the task.
We define RC as:
\begin{equation}
    \mathcal{L}_{\text{RC}}(x_S, \Tilde{x}_S) = \left( f_{\text{RC}}(x_S, \Tilde{x}_S;\theta_S) - f_{\text{RC}}(x_S;\theta_{IN}) \right)^2,
\end{equation}
where $f_{\text{RC}}(x_S, \Tilde{x}_S;\theta_S)$ denotes the bottleneck feature of original sample $x_S$ and style-diversified sample $\Tilde{x}_S$, which are obtained from the ongoing model $\theta_S$. $f(x_S;\theta_{IN})$ denotes the bottleneck feature of original sample $x_S$ in the retrospective ImageNet pre-trained model $\theta_{IN}$.

\noindent\textbf{Discussion.} Our retrospection consistency is inspired by the Feature Distance (FD) in DAFormer~\citep{hoyer2021daformer} but with different motivation and implementation. \textit{First}, DAFormer focuses on unsupervised domain adaptation in semantic segmentation with unlabeled real-world data available, so it only focuses on fitting to the specific target domain (CityScapes~\citep{cityscapes}) rather than addressing unseen domain shift. \textit{Second}, FD in DAFormer is used to better classify those similar classes (\eg, bus and train) with the classification knowledge from ImageNet. %
In this paper, we focus on the domain generalization in visual recognition tasks.
Since we have no idea about the target distribution, RC in our framework serves as an important guidance for the general visual knowledge, leading the model less overfitting to the source data. In addition, apart from semantic segmentation, RC can be applied to image classification and object detection.

\subsection{Applications of Dual Consistency Learning}
The recognition objective is different for each task: image classification leverages the global information to classify the whole image; semantic segmentation is required to classify each pixel to the pre-defined categories; object detection aims to detect the object and recognize the class. Consequently, the implementation of dual consistency constraints is based on the task requirements.

\noindent\textbf{Image Classification.}
The input image $x$ first forwards through the backbone $\theta$ and the global average pooling layer to obtain the backbone feature $f(x;\theta)$, which is used for retrospective consistency. Then a linear classifier is used to predict the image class for backbone features of both the original and stylized samples. The prediction after softmax is the posterior probability $P$ and $\Tilde{P}$.

\noindent\textbf{Semantic Segmentation.}
Since the pixel-level prediction is required for semantic segmentation, the consistency is applied to each pixel instead of the whole image.
In addition, semantic segmentation in the urban scene contains both foreground (\eg, car and person) and background (\eg, road and wall) classes, while the ImageNet pre-trained model only focuses on the foreground categories. Thus, only the foreground backbone features are used for retrospective consistency. Here we define the mask of foreground pixels as $M_{fg}$ and the retrospective features $f_{\text{RC}}(x;\theta)$ is $M\cdot f(x;\theta)$. 
Nevertheless, since the style variation is applied to both foreground and background pixels, so the style consistency is applied to the posterior probabilities of all the pixels, which are the predicted value after classifier and softmax.

\noindent\textbf{Object Detection.}
Object detection requires to classify the object category within the bounding box instead of the whole image or each pixel. Thus, the consistency should be applied only to the box features. Specifically, we use the ground truth box to obtain the RoI (Region of Interest) features from the backbone features via RoI align~\citep{he2017mask}, and the RoI features $f_{\text{RoI}}(x)$ serve as the retrospective features $f_{\text{RC}}(x;\theta)$. After that, the RoI features are fed into the classifier to obtain the posterior probability $P$ and $\Tilde{P}$ for each object, which are used in the style consistency.

\begin{figure}[t]
    \centering
    \includegraphics[width=.9\linewidth]{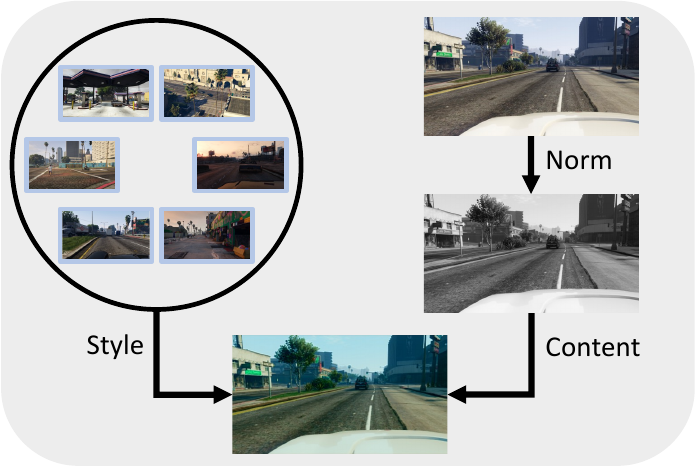}
    \caption{Illustration of the style hallucination module}
    \label{fig:shm}
\end{figure}

\subsection{Style Hallucination}
\label{sec:sh}
\noindent\textbf{Background.} Style transfer~\citep{chen2021diverse,adain} and domain generalization~\citep{crossnorm,zhou2021mixstyle} methods show that the channel-wise mean and standard deviation can represent the non-content style of the image, which plays an important role in the domain shift. The style features can be readily used by AdaIN~\citep{adain} which can transfer the image to an arbitrary style while remaining the content:
\begin{equation}
    \text{AdaIN}(x, y)=\sigma(y)\left(\frac{x-\mu(x)}{\sigma(x)}\right)+\mu(y),
\end{equation}
where $x$ and $y$ denotes the feature maps providing the content and style respectively. $\mu(*)$ and $\sigma(*)$ denotes the channel-wise mean and standard deviation. In domain generalization, as only one or multiple source domains are accessible, previous works modify AdaIN by replacing style features with other source styles. Those styles can be directly obtained from other samples~\citep{crossnorm} or can be generated by mixing other styles with its own styles~\citep{zhou2021mixstyle}.

\noindent\textbf{Style Hallucination Module (SHM).}
The ways of previous methods~\citep{crossnorm,zhou2021mixstyle} in generating extra styles are sub-optimal, since they just randomly swap or mix source styles without considering the frequency and diversity of the source styles. As a result, more samples of the dominant style (\eg, daytime) will be generated, yet the generated distribution may be quite different from the target one. 
Since we have no idea about the distribution of target set, it is better to diversify the source samples as much as possible. We next introduce the Style Hallucination Module (SHM) for generating diverse source samples. The illustration of SHM is shown in Fig.~\ref{fig:shm}.

\begin{figure*}[ht]
    \centering
    \includegraphics[width=.8\linewidth]{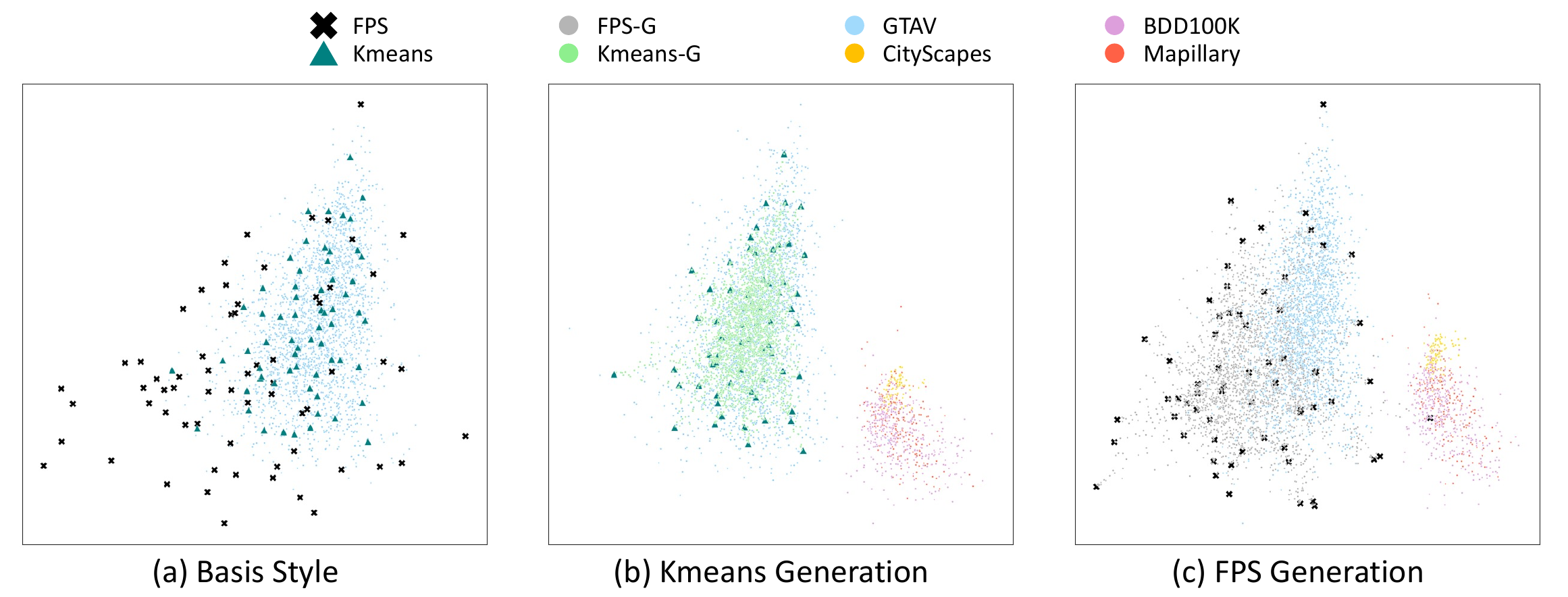}
    \caption{Visualization of distributions of different domains. (a) Comparison of two kinds of basis styles; (b) generated style with Kmeans basis style; (c) generated style with FPS basis style. (Zoom in for details.) }
    \label{fig:visual-dist}
\end{figure*}

\noindent\textbf{Definition 1:} 
\textit{A basis $B$ of a vector space $V$ over a field $F$ is a linearly independent subset of $V$ that spans $V$. When the field is the reals $\mathbb{R}$, the resulting basis vectors are $n$-tuples of reals that span $n$-dimensional Euclidean space $\mathbb{R}^n$}~\citep{basis}.

According to Definition 1, style space can be viewed as a subspace of $C$-dimensional vector space, and thus all possible styles can be represented by the basis vectors.
However, if we directly take $C$ linearly independent vectors as the basis, \eg, orthogonal unit vectors, many unrealistic styles %
are generated since the realistic styles are only in a small subspace, and such generated styles %
can impair the model training. 
To reconcile the diversity and realism, we use farthest point sampling~(FPS)~\citep{qi2017pointnet++} to select $C$ styles from all the source styles as basis styles. 
FPS is widely used for point cloud downsampling, which can iteratively choose $C$ points from all the points, such that the chosen points are the most distant points with respect to the remaining points.
Despite not strictly linearly independent, basis styles obtained by FPS can represent the style space to the utmost extent, and also contain many rare styles since rare styles are commonly far away from dominant ones. 
In addition, we recalculate the dynamic basis styles every $k$ epochs instead of fixing them, as the style space is changing along with the model training. To generate new styles, we sample the combination weight $W=[w_1, \cdots, w_C]$ from Dirichlet distribution $B([\alpha_1, \cdots, \alpha_C])$ with the concentration parameters $[\alpha_1, \cdots, \alpha_C]$ all set to $1/C$. 
The basis styles are then linearly combined by $W$:
\begin{equation}
    \mu_{HS} = W\cdot \mu_{base}, \qquad \sigma_{HS} = W \cdot \sigma_{base},
\end{equation}
where $\mu_{base}\in \mathbb{R}^{C\times C}$ and $\sigma_{base} \in \mathbb{R}^{C\times C}$ are the $C$ basis styles. With the generated styles, style hallucinated samples $\Tilde{x}_S$ can be obtained by:
\begin{equation}
\centering
    \Tilde{x}_S = \sigma_{HS}\left(\frac{x_S-\mu(x_S)}{\sigma(x_S)}\right)+\mu_{HS}.
\end{equation}

\noindent\textbf{Discussion.}
We take the semantic segmentation benchmark as an example to verify the FPS basis style selection. In Fig.~\ref{fig:visual-dist}, GTAV~\citep{gtav} is the source domain while CityScapes~\citep{cityscapes}, BDD100K~\citep{bdd} and Mapillary~\citep{mapillary} are the target domains.
Selecting representative basis styles is crucial for SHM. FPS is adopted in our method as it can cover the rare styles to the utmost extent. Another way is taking the Kmeans~\citep{macqueen1967some_kmeans} clustering centers as the basis. 
As shown in Fig.~\ref{fig:visual-dist}(a), FPS samples (\textbf{\textcolor{black}{black cross}}) spread out more than Kmeans centers (\textbf{\textcolor{teal}{teal triangle}}), and can cover almost all possible source styles (\textbf{\textcolor{lightskyblue}{lightskyblue point}}). When using the basis styles for style generation, styles obtained from Kmeans centers (Fig.~\ref{fig:visual-dist}(b)) are still within the source distribution and even ignore some possible rare styles. In contrast, FPS basis styles can generate more diverse styles (Fig.~\ref{fig:visual-dist}(c)), and even generate some styles close to the real-world ones (\textbf{\textcolor{yellow}{yellow}, \textcolor{pink}{pink} and \textcolor{orange}{orange} point}). Tab.~\ref{table:style-variation} further demonstrates the effectiveness of FPS basis styles and shows that the Kmeans basis styles are even worse than directly swapping and mixing source styles.

\subsection{Training Objective}
The overall training objective is the combination of the task loss and the proposed two consistency constraints:
\begin{equation}
\small
\begin{aligned}
    \mathcal{L} &= \mathcal{L}_{\text{task}}(x_S, y_S) + \lambda_{\text{SC}} \mathcal{L}_{\text{SC}}(x_S, \Tilde{x}_S) + \lambda_{\text{RC}} \mathcal{L}_{\text{RC}}(x_S, \Tilde{x}_S),
\end{aligned}
\end{equation}
where $\lambda_{\text{SC}}$ and $\lambda_{\text{RC}}$ are the weights for style consistency and retrospection consistency, respectively. The task loss is the cross entropy loss for image classification and semantic segmentation, and is the RPN loss together with cross entropy loss for object detection.

\section{Experiments}
\label{sec:exp}
In this section, we conduct experiments on image classification, semantic segmentation and object detection to demonstrate the superiority of \ours. 
We first compare our method with state-of-the-art methods in the three visual tasks. Then, we provide ablation studies on image classification and semantic segmentation to verify the effectiveness of each proposed component. Finally, we analyze and evaluate \ours on semantic segmentation and provide the visualization results. 

\subsection{Semantic Segmentation}
\label{sec:exp-seg}
\subsubsection{Experimental Setup}

\noindent\textbf{Datasets.}
Two synthetic datasets (GTAV~\citep{gtav} and SYNTHIA~\citep{synthia}) and three real-world datasets (CityScapes~\citep{cityscapes}, BDD100K~\citep{bdd} and Mapillary~\citep{mapillary}) are used in this paper.
GTAV~\citep{gtav} contains 24,966 images with the size of 1914$\times$1052, splitting into 12,403 training, 6,382 validation, and 6,181 testing images. SYNTHIA~\citep{synthia} contains 9,400 images of 960$\times$720, where 6,580 images are used for training and 2,820 images for validation.
CityScapes~\citep{cityscapes} contains 2,975 training images and 500 validation images of 2048$\times$1024. BDD100K~\citep{bdd} and Mapillary~\citep{mapillary} contain 7,000 and 18,000 images for training, and 1,000 and 2,000 images for validation, respectively.

\noindent\textbf{Implementation Details.} 
We use both ConvNets and Transformer backbone to conduct experiments. For the \textit{ConvNets models}, we use DeepLabV3+~\citep{deeplab} as the segmentation model and equip it with two backbones: ResNet-50 and ResNet-101~\citep{he2016deep}. The SHM is inserted after the first Conv-BN-ReLU layer (layer0). We re-select the basis styles with the interval $k=3$. We set $\lambda_{\text{SC}}=10$ and $\lambda_{\text{RC}}=1$.
Models are optimized by the SGD optimizer with the learning rate 0.01, momentum 0.9 and weight decay 5$\times$10$^{-4}$. The polynomial decay~\citep{liu2015parsenet} with the power of 0.9 is used as the learning rate scheduler. All models are trained with the batch size of 8 for 40K iterations. 
During training, four widely used data augmentation techniques are adopted, including color jittering, Gaussian blur, random flipping, and random cropping of 768$\times$768. 
For the \textit{Transformer models}, we use SegFormer~\citep{xie2021segformer} with MiT-B5~\citep{xie2021segformer} as the segmentation model. SHM is inserted after the first block and the basis styles are re-selected every 4k iterations. We set $\lambda_{\text{SC}}=10$ and $\lambda_{\text{RC}}=0.005$.
Following \cite{hoyer2021daformer}, models are optimized by the AdamW~\citep{adamw} optimizer with the learning rate 6$\times$10$^{-5}$ for encoder and 6$\times$10$^{-4}$ for decoder, and weight decay 0.01. 
Linear learning rate warmup by 1.5k iterations is first adopted, and then the learning rate linearly decays.
All models are trained with the batch size of 2 for 40K iterations. 

\noindent\textbf{Protocols.}
In this section, we focus on three protocols. (1) \textit{Synthetic-to-real single-source DG} takes the GTAV as the source domain and evaluates the model on the three real-world target domains. To conduct a fair comparison with \cite{robustnet} and \cite{FSDR}, we train the model with GTAV training data (12,403 images) when using ResNet-50 backbone (same with \cite{robustnet}), and with the whole GTAV datasets (24,966 images) when using ResNet-101 backbone (same with \cite{FSDR}) and MiT-B5 backbone. 
(2) \textit{Synthetic-to-real multi-source DG} takes the GTAV and SYNTHIA as the source domains and evaluate on the three real-world target domains.
We follow \cite{robustnet} to train the model with the training set of GTAV (12,403 images) and SYNTHIA (6,580 images) using ResNet-50 backbone.
(3) \textit{Real-to-others single-source DG} is also investigated in this paper. Following \cite{robustnet}, we train the model with the training set of CityScapes and evaluate it on the other four synthetic and real-world datasets.

\noindent\textbf{Evaluation Metric.} 
We use the 19 shared semantic categories for training and evaluation. The mean intersection-over-union (mIoU) of the 19 categories on the target datasets is adopted as the evaluation metric.

\begin{table*}[t]
\centering
\footnotesize
\setlength{\tabcolsep}{3pt}
\begin{tabular}{l|l|c|c|p{1.6cm}<\centering|p{1.6cm}<\centering|p{1.6cm}<\centering|p{1.6cm}<\centering}
\toprule
Net & Methods (GTAV) & Venue & Extra Data & CityScapes& BDD100K & Mapillary& Avg. \\
\midrule
\multicolumn{1}{c|}{\multirow{10}{*}{\rotatebox{90}{\textbf{ResNet-50}}}} & Baseline& --- &\xmark & 28.95 & 25.14& 28.18& 27.42\\
\cmidrule{2-8}
&IBN-Net~\citep{ibn} & ECCV 18 &\xmark & 33.85 & 32.30 & 37.75& 34.63\\
&IterNorm~\citep{huang2019iterative} & CVPR 19 &\xmark & 31.81 & 32.70 & 33.88 & 32.79 \\
&SW~\citep{pan2019switchable} & ICCV 19 &\xmark &29.91 & 27.48 & 29.71 & 29.03 \\
&DRPC~\citep{DRPC}$^{\S}$ & ICCV 19 &\textcolor{red}{\cmark} & 37.42 & 32.14& 34.12& 34.56\\
&ISW~\citep{robustnet} & CVPR 21 & \xmark & 36.58 & 35.20 & 40.33& 37.37 \\
&GTR~\citep{gtr} & TIP 21 &\xmark & 37.53 & 33.75 & 34.52 & 35.27 \\
&SAN-SAW~\citep{san} & CVPR 22&\xmark &39.75 & 37.34 & 41.86 & 39.65 \\
&AdvStyle~\citep{advstyle}& NeurIPS 22 &\xmark & 39.60 & 38.59 & 41.89 & 40.03 \\
& \bf Ours & --- & \xmark & \textbf{44.65} & \textbf{39.28} & \textbf{43.34} & \textbf{42.42} \\
\midrule
\multicolumn{1}{c|}{\multirow{9}{*}{\rotatebox{90}{\textbf{ResNet-101}}}} & Baseline & --- &\xmark & 32.97& 30.77& 30.68& 31.47\\
\cmidrule{2-8}
& IBN-Net~\citep{ibn}  & ECCV 18 &\xmark & 37.37& 34.21& 36.81& 36.13 \\
& DRPC~\citep{DRPC}$^{\S}$ & ICCV 19 &\textcolor{red}{\cmark}   & {42.53} & {38.72} & {38.05} & 39.77\\
& ISW~\citep{robustnet}$^\dagger$  & CVPR 21&\xmark & 37.20 & 33.36& 35.57& 35.38\\
& FSDR~\citep{FSDR}$^{\S}$ & CVPR 21&\textcolor{red}{\cmark}  & {44.80}  & 41.20 & 43.40 & 43.13\\
&GTR~\citep{gtr} & TIP 21&\xmark &43.70& 39.60 & 39.10 & 40.80\\
&SAN-SAW~\citep{san} & CVPR 22 &\xmark &45.33 &41.18 & 40.77 &42.43 \\
&AdvStyle~\citep{advstyle}& NeurIPS 22&\xmark & 44.51 & 39.27 & 43.48 & 42.42 \\
& \bf Ours & --- &\xmark & \textbf{46.66} & \textbf{43.66} & \textbf{45.50} & \textbf{45.27} \\
\midrule
\multicolumn{1}{c|}{\multirow{5}{*}{\rotatebox{90}{\textbf{MiT-B5}}}} & Baseline & --- & \xmark & 45.60 & 44.86 & 48.72 & 46.39 \\
\cmidrule{2-8}
& MixStyle~\citep{zhou2021mixstyle}$^\dagger$ & ICLR 21&\xmark & 43.29&44.23&46.51& 44.68 \\
& CrossNorm~\citep{crossnorm}$^\dagger$ & ICCV 21&\xmark & 46.41 & 44.69 & 50.21 & 47.10 \\
& AdvStyle~\citep{advstyle}$^\dagger$ & NeurIPS 22 &\xmark & 46.56 &45.10&48.35&46.67 \\
& \bf Ours  & --- & \xmark & \bf 53.27 & \bf 48.19 & \bf 54.99 & \bf 52.15 \\

\bottomrule
\end{tabular}
\caption{Comparison with state-of-the-art methods on single-source DG-Seg with ResNet-50, ResNet-101, and MiT-B5 backbones, respectively. ``Extra Data'' denotes using extra real-world data during training. ${\S}$ denotes selecting best checkpoint for each target dataset, and $\dagger$ denotes reproducing the results based on the source code.}
\label{tab:gtav-sota}
\end{table*}

\begin{table*}[t]
\centering
\footnotesize
\setlength{\tabcolsep}{3pt}
\begin{tabular}{l|c|p{2cm}<\centering|p{2cm}<\centering|p{2cm}<\centering|p{2cm}<\centering}
\toprule
Methods (G+S) & Venue & CityScapes& BDD100K & Mapillary& Avg. \\
\midrule
Baseline & --- & 35.46& 25.09  & 31.94  & 30.83\\
\midrule
IBN-Net~\citep{ibn} & ECCV 18 & 35.55& 32.18  & 38.09  & 35.27\\
MLDG~\citep{li2018learning} & AAAI 18 & 38.84 & 31.95 & 35.60 & 35.46 \\
ISW~\citep{robustnet} & CVPR 21& 37.69& 34.09  & 38.49  & 36.76\\
PTM~\citep{kim2022pin}& CVPR 22 & 44.51 &38.07 & 42.70 &41.76 \\
AdvStyle~\citep{advstyle}& NeurIPS 22 & 39.29 & 39.26 & 41.14 & 39.90 \\
\bf Ours  & --- &\textbf{47.43} & \textbf{40.30} & \textbf{47.60} & \textbf{45.11} \\
\bottomrule
\end{tabular}
\caption{Comparison with state-of-the-art methods on multi-source DG-Seg. All models use ResNet-50 backbone and are trained with training sets of GTAV and SYNTHIA.}
\label{tab:multi-src}
\end{table*}

\begin{table*}[t]
\centering

\footnotesize
    \centering
    \begin{tabular}{l|p{1.8cm}<\centering|p{1.8cm}<\centering|p{1.8cm}<\centering|p{1.8cm}<\centering|p{1.8cm}<\centering}
    \toprule
      Methods & GTAV & SYNTHIA & BDD100K & Mapillary & Avg. \\
    \midrule
    Baseline & 42.55 & 23.29 & 44.96 & 51.68 & 40.62 \\
    \midrule
    SW~\citep{pan2019switchable} & 44.87 & 26.10 & 48.49 & 55.82 & 43.82 \\
    IterNorm~\citep{huang2019iterative} & 45.73 & 25.98 & 49.23 & 56.26 & 44.30 \\
    IBN-Net~\citep{ibn} & 45.06 & 26.14 & 48.56 & 57.04 & 44.20 \\
    ISW~\citep{robustnet}  & 45.00 & 26.20 & 50.73 & 58.64 & 45.14\\
    \textbf{Ours}  & \textbf{48.61} & \textbf{27.62} & \textbf{50.95} & \textbf{60.67} & \bf 46.96 \\
    \bottomrule
    \end{tabular}
\caption{Comparison with state-of-the-art methods on Real-to-Others DG-Seg. All models use ResNet-50 backbone and are trained with GTAV training set.}
\label{tab:city}
\end{table*}

\subsubsection{Comparison with State-of-the-art Methods}

\noindent\textbf{Synthetic-to-Real Single-Source DG.}
In Tab.~\ref{tab:gtav-sota}, we compare \ours with state-of-the-art methods under single-source setting, including SW~\citep{pan2019switchable}, IterNorm~\citep{huang2019iterative}, IBN-Net~\citep{ibn}, ISW~\citep{robustnet}, DRPC~\citep{DRPC}, FSDR~\citep{FSDR}, GTR~\citep{gtr}, SAN-SAW~\citep{san} and AdvStyle~\citep{advstyle}.
\textit{First}, we compare models that are trained with GTAV training set, using ResNet-50 backbone. \ours achieves an average mIoU of 42.42\% on the three real-world target datasets, yielding an improvement of 15.00\% mIoU over the baseline and outperforming the previous best method (AdvStyle) by 2.39\%.
\textit{Second,} we further compare methods under the training protocol of DRPC~\citep{DRPC} and FSDR~\citep{FSDR}, using ResNet-101 backbone and taking the whole set of GTAV (24,966 images) as the training data. 
Note that DRPC~\citep{DRPC} and FSDR~\citep{FSDR} utilize extra real-world data from ImageNet~\citep{imagenet} or even driving scenes. Moreover, they select the best checkpoint for each target dataset, which is impractical in the real-world applications. 
Even so, we achieve the best results on all three datasets, \textbf{46.66\% on CityScapes, 43.66\% on BDD100K, and 45.50\% on Mapillary}, outperforming FSDR~\citep{FSDR} and AdvStyle~\citep{advstyle} by by 2.14\% and 2.85\% in the average mIoU. 
\textit{Third}, we further investigate \ours with the Transformer backbone. Since no previous work leveraging the Transformer backbone, we reproduce MixStyle~\citep{zhou2021mixstyle}, CrossNorm~\citep{crossnorm} and AdvStyle~\citep{advstyle} based on the source code.
As shown in Tab.~\ref{tab:gtav-sota}, the source only model serves as a strong baseline and these method can hardly improve the performance. Instead, \ours outperforms the baseline by \textbf{7.67\% on CityScapes, 3.33\% on BDD100K, and 6.27\% on Mapillary}, respectively.
These results show that we produce new state of the art in domain generalized semantic segmentation with different segmentation models and backbones.

\noindent\textbf{Synthetic-to-Real Multi-Source DG.} To further verify the effectiveness of \ours, we compare \ours with IBN-Net~\citep{ibn}, ISW~\citep{robustnet}, PTM~\citep{kim2022pin} and AdvStyle~\citep{advstyle} under the multi-source setting.
We use ResNet-50 as the backbone and take the training set of GTAV and SYNTHIA as the source domains. As shown in Tab.~\ref{tab:multi-src}, \ours gains an improvement of 14.28\% in average mIoU over the baseline, and outperforms PTM~\citep{kim2022pin} and AdvStyle~\citep{advstyle} by 3.35\% and 5.21\% respectively. 
The significant improvement over previous works is mainly benefited from the various samples. With richer source samples, our SHM can generate more informative and diverse styles, which can effectively facilitate the dual consistency learning.

\noindent\textbf{Real-to-Others Single-Source DG.}
In Tab.~\ref{tab:city}, we compare \ours with previous methods under the real-to-others DG setting, where CityScapes~\citep{cityscapes} is leveraged as the source domain and the model is generalized to real (BDD100K~\citep{bdd} and Mapillary~\citep{mapillary}) and synthetic (GTAV~\citep{gtav} and SYNTHIA~\citep{synthia}) domains.
As shown in Tab.~\ref{tab:city}, \ours consistently outperforms ISW~\citep{robustnet} and IBN-Net~\citep{ibn} on both real and synthetic datasets. These results further verify the versatility of our method.

\begin{table}[t]
\footnotesize
    \setlength{\tabcolsep}{2.8pt}
    \centering
    \begin{tabular}{l|cccc|c}
    \toprule
         Method & Art. & Car.& Ske. & Pho. & Avg.  \\
         \midrule
         ERM & 67.4 & 74.4 &51.4&42.6&58.9\\
         JiGen~\citep{carlucci2019domain} & 69.1& 74.6 & 52.4& 41.5&59.4\\
         RSC~\citep{huangRSC2020} & 68.8 &74.5 &53.6 &41.9 & 59.7\\
         L2D~\citep{wang2021L2D} &74.3&77.5& 54.4 & 45.9 & 63.0\\
         \midrule
         \bf ERM+SHADE & \bf 76.5 & 77.2 & \bf 60.1 & 49.3 & 65.8 \\
         \bf RSC+SHADE & 75.4 & 74.8 & 55.8 & 49.8 & 64.0 \\
         \bf L2D+SHADE & 76.1 & \bf 79.7 & \bf 60.1 & \bf 51.0 & \bf 66.7 \\
    \bottomrule
    \end{tabular}
    \caption{Comparison with state-of-the-art methods on PACS image classification benchmark. One domain (name in column) is used as the source (training) data and the other domains are used as the target (testing) data.}
    \label{tab:pacs}
\end{table}

\subsection{Image Classification}
\subsubsection{Experimental Setup}
\noindent\textbf{Datasets.}
We conduct experiments on the PACS~\citep{pacs} benchmark. 
{PACS}~\citep{pacs} contains four domains (Artpaint, Cartoon, Sketch, and Photo), and each domain contains 224 $\times$ 224 images belonging to seven categories. There are 9,991 images in total. 

\noindent\textbf{Implementation Details.}
Following \cite{huangRSC2020}, we use the ResNet18~\citep{he2016deep} pretrained on ImageNet~\citep{imagenet} as the backbone.
The SHM is inserted after the first Conv-BN-ReLU layer and the basis styles are selected every 3 epochs. The loss weight $\lambda_{\text{SC}}$ and $\lambda_{\text{RC}}$ are set to 10 and 0.1, respectively.
We train the model by SGD optimizer. The learning rate is initially set to 0.004 and divided by 10 after 24 epochs. The model is trained for 30 epochs in total with a batch size of 128.

\noindent\textbf{Evaluation Metric.} 
We select one of the four domains in PACS as the source domain and the other domains as the target domains. 
The mean accuracy of the 7 categories on the target datasets is adopted as the evaluation metric.

\subsubsection{Comparison with State-of-the-art Methods}

We compare \ours with the baseline (ERM~\citep{vapnik2013nature}) and three state-of-the-art DG methods, including JiGen~\citep{carlucci2019domain}, RSC~\citep{huangRSC2020} and L2D~\citep{wang2021L2D}. We reproduce JiGen~\citep{carlucci2019domain}, RSC~\citep{huangRSC2020} and L2D~\citep{wang2021L2D} with their official source codes. All methods use the same baseline (ERM~\citep{vapnik2013nature}). 
To verify the effectiveness of \ours, we apply \ours to ERM~\citep{vapnik2013nature}, RSC~\citep{huangRSC2020} and L2D~\citep{wang2021L2D}.
As shown in Tab.~\ref{tab:pacs}, combining \ours with ERM~\citep{vapnik2013nature} can outperform state-of-the-art method (L2D~\citep{wang2021L2D}) on all four domains by 2.2\%, 0.3\%, 5.7\%, and 3.4\%, respectively.
In addition, \ours can yield improvements on all the three models by 6.9\% on ERM~\citep{vapnik2013nature}, 4.3\% on RSC~\citep{huangRSC2020}, and 3.7\% on L2D~\citep{wang2021L2D} in the average accuracy.
The above results demonstrate \ours is versatile and effective.

\begin{table*}[ht]
\footnotesize
    \setlength{\tabcolsep}{2.8pt}
    \centering
    \begin{tabular}{l|p{1.8cm}<\centering| p{1.8cm}<\centering| p{1.8cm}<\centering| p{1.8cm}<\centering | p{1.8cm}<\centering}
    \toprule
         Method & Night Sunny & Dusk Rainy & Night Rainy & Day Foggy & Avg.  \\
         \midrule
         Baseline & 33.5 &  26.6 & 14.5 & 31.9 & 26.6 \\
        \midrule
        SW~\citep{pan2019switchable} & 33.4 & 26.3 & 13.7 & 30.8 & 26.1 \\
        IBN-Net~\citep{ibn} & 32.1 & 26.1 & 14.3 & 29.6 & 25.5\\
        IterNorm~\citep{huang2019iterative} & 29.6 & 22.8 & 12.6 & 28.4 & 23.4 \\
        ISW~\citep{robustnet} & 33.2 & 25.9 & 14.1 & 31.8 & 26.3 \\
        Single-DGOD~\citep{wu2022single} &  \bf 36.6 & 28.2 & 16.6 & \bf 33.5 & \bf 28.7 \\
         \bf Ours & 33.9 & \bf 29.5 & \bf 16.8 & 33.4 & 28.4 \\
    \bottomrule
    \end{tabular}
    \caption{Comparison with state-of-the-art methods on domain generalized object detection benchmark.}
    \label{tab:detection}
\end{table*}

\begin{table*}[t]
\begin{center}
\footnotesize
\setlength{\tabcolsep}{3pt}
\begin{tabular}{c|p{0.9cm}<\centering|p{0.9cm}<\centering|p{0.9cm}<\centering|p{0.9cm}<\centering|p{1.6cm}<\centering|p{1.6cm}<\centering|p{1.6cm}<\centering|p{1.6cm}<\centering}
\toprule
No. & SHM & $\mathcal{L}_{\text{SC}}$ & $\mathcal{L}_{\text{RC}}$ & EMA & CityScapes & BDD100K & Mapillary & Avg. \\
\midrule
1 & \xmark & \xmark & \xmark & \xmark & 28.95 & 25.14 & 28.18 & 27.42\\
\midrule
2 &\textcolor{dark-green}{\cmark} & \xmark & \xmark & \xmark & 38.68	& 32.40 & 35.96 & 35.68 \\
3&\textcolor{dark-green}{\cmark} & \textcolor{dark-green}{\cmark} & \xmark & \xmark & 42.66 & 35.92 & 40.42 & 39.67 \\
4&\textcolor{dark-green}{\cmark} & \xmark & \textcolor{dark-green}{\cmark} & \xmark & 41.43 & 37.65 & 41.77 & 40.29 \\
5&\textcolor{dark-green}{\cmark} & \textcolor{dark-green}{\cmark} & \xmark & \textcolor{dark-green}{\cmark} &42.38 & 38.04 & 42.34 & 40.92 \\
6&\textcolor{dark-green}{\cmark} & \textcolor{dark-green}{\cmark} & \textcolor{dark-green}{\cmark} & \xmark & \textbf{44.65} & \textbf{39.28} & \textbf{43.34} & \textbf{42.42} \\
\bottomrule
\end{tabular}
\caption{Ablation studies of each component on the synthetic-to-real single source DG benchmark. All models use ResNet-50 backbone and are trained with GTAV training set. SHM: style hallucination module; EMA: using exponential moving average model instead of ImageNet pre-trained model.}
\label{table:ablation-loss}
\end{center}
\end{table*}

\begin{table*}[t]
\begin{center}
\footnotesize
\setlength{\tabcolsep}{3pt}
\begin{tabular}{c|p{1.1cm}<\centering|p{1.1cm}<\centering|p{1.1cm}<\centering|p{1.6cm}<\centering|p{1.6cm}<\centering|p{1.6cm}<\centering|p{1.6cm}<\centering|p{1.6cm}<\centering}
\toprule
No.&SHM & $\mathcal{L}_{\text{SC}}$ & $\mathcal{L}_{\text{RC}}$ & Art. & Car. & Ske. & Pho. & Avg. \\
\midrule
1&\xmark & \xmark & \xmark & 67.4 & 74.4 & 51.4 & 42.6 & 58.9 \\
\midrule
2&\textcolor{dark-green}{\cmark} & \xmark & \xmark & 69.9 & 73.9 & 51.1 &43.4 & 59.6 \\
3&\textcolor{dark-green}{\cmark} & \textcolor{dark-green}{\cmark} & \xmark & 72.8 & 74.8 & 60.0 & 49.1 & 64.2 \\
4&\textcolor{dark-green}{\cmark} & \xmark & \textcolor{dark-green}{\cmark} & 69.6 & 75.8 & 54.3 & 43.5 & 60.8\\
5&\textcolor{dark-green}{\cmark} & \textcolor{dark-green}{\cmark} & \textcolor{dark-green}{\cmark} &  \bf 76.5 & \bf 77.2 & \bf 60.1 & \bf 49.3 & \bf 65.8 \\

\bottomrule
\end{tabular}
\caption{Ablation studies of each component on the PACS benchmark. ERM baseline with ResNet-18 backbone is used. SHM: style hallucination module.}
\label{table:ablation-classification}
\end{center}
\end{table*}

\begin{table*}[t]
\begin{center}
\footnotesize
\setlength{\tabcolsep}{3pt}
\begin{tabular}{l|p{2cm}<\centering|p{2cm}<\centering|p{2cm}<\centering|p{2cm}<\centering}
\toprule
Methods (GTAV) & CityScapes & BDD100K & Mapillary & Avg. \\
\midrule
Baseline & 28.95 & 25.14 & 28.18 & 27.42\\
\midrule
Random Style & 37.99 & 37.63 & 38.06 &	37.89 \\
MixStyle~\citep{zhou2021mixstyle} & 43.14 & 37.94 & 42.22 & 41.10 \\
CrossNorm~\citep{crossnorm} & 43.13	&	37.20	&	41.83	& 40.72 \\
Kmeans Basis & 40.50 & 37.62 & 39.46 & 39.19 \\
\bf Ours & \textbf{44.65} & \textbf{39.28} & \textbf{43.34} & \textbf{42.42} \\
\bottomrule
\end{tabular}
\caption{Comparison of different style variation methods. All models use ResNet-50 backbone and are trained with GTAV training set.}
\label{table:style-variation}
\end{center}
\end{table*}

\subsection{Object Detection}
\subsubsection{Experimental Setup}
\noindent\textbf{Datasets.}
We use the urban-scene detection benchmark~\citep{wu2022single} in this paper. The benchmark includes five datasets of different weather conditions. Specifically, daytime-sunny dataset contains 19,395 and 8,313 images from BDD100K~\citep{bdd} for training and testing, respectively.
The night-sunny dataset contains 26,158 images from BDD100K. The dusk-rainy and night-rainy datasets include 3,501 and 2,494 images rendered from BDD100K by \cite{wu2021vector}.
In addition, 3,775 foggy images from FoggyCityscapes~\citep{sakaridis2018semantic} and Adverse-Weather~\citep{hassaballah2020vehicle} form the daytime-foggy dataset.

\noindent\textbf{Implementation Details.}
Following \cite{wu2022single}, we use Faster R-CNN~\citep{ren2015faster} with ResNet-101~\citep{he2016deep} backbone as the detector. 
The SHM is inserted after the first Conv-BN-ReLU layer and the basis styles are selected every epoch. The loss weight $\lambda_{\text{SC}}$ and $\lambda_{\text{RC}}$ are set to 10 and 0.1, respectively.
SGD optimizer is used to train the model with learning rate 0.004, momentum 0.9 and weight decay 5$\times$10$^{-4}$. 
The model is trained for 10 epochs in total with a batch size of 2.

\noindent\textbf{Evaluation Metric.} 
The daytime-sunny dataset is used as the source training data, while the night-sunny, dusk-rainy, night-rainy and daytime-foggy datasets are used for the target testing data.
We use the 7 shared categories for training and evaluation and the mean average precision (mAP) with a threshold of 0.5 is adopted as the evaluation metric.

\subsubsection{Comparison with State-of-the-art Methods}
We compare \ours with the baseline model, SW~\citep{pan2019switchable}, IBN-Net~\citep{ibn}, IterNorm~\citep{huang2019iterative}, ISW~\citep{robustnet} and Single-DGOD~\citep{wu2022single} in Tab.~\ref{tab:detection}. The results are borrowed from \cite{wu2022single}.
As shown in Tab.~\ref{tab:detection}, \ours outperforms the baseline model on all the target datasets by 0.4\% on night-sunny, 2.9\% on dusk-rainy, 2.3\% on night-rainy, and 1.5\% on daytime-foggy. Furthermore, \ours achieves the state-of-the-art performance on dusk-rainy and night-rainy datasets. These results demonstrate that \ours is applicable to improving the generalization ability for the object detection model.

\subsection{Ablation Studies}
To investigate the effectiveness of each component in \ours, we conduct ablation studies in Tab.~\ref{table:ablation-loss} and Tab.~\ref{table:ablation-classification} on semantic segmentation and image classification, respectively.

\noindent\textbf{Effectiveness of Style Hallucination Module (SHM).}
SHM is the basis of \ours. When using SHM only, we directly apply cross entropy loss on the style hallucinated samples. As shown in the second row of Tab.~\ref{table:ablation-loss}, our SHM can largely improves the model performance even without using the proposed dual consistency learning. This demonstrates the importance of training the model with diverse samples and the effectiveness of our SHM.

\noindent\textbf{Effectiveness of Style Consistency (SC).}
SC is the consistency constraint that leads the model to learn style invariant representation. In Tab.~\ref{table:ablation-loss}, compared with only applying SHM, SC yields an improvement of 3.99\% in average mIoU, demonstrating the superiority of the proposed logit pairing over cross entropy loss in learning style invariant model.

\noindent\textbf{Effectiveness of Retrospection Consistency (RC).}
\textit{First}, RC serves as an important guidance to ameliorate the overfitting problem
by general visual knowledge. 
Applying RC on top of SHM can yield an improvement of 4.61\%,
while removing RC will degrade the performance of \ours by 2.75\% in mIoU. 
\textit{Second}, we conduct experiments to verify that the effectiveness of RC lies in the general visual knowledge instead of feature-level distance minimization of the paired samples. 
As directly minimizing the feature-level absolute distance of paired samples will lead to sub-optimal results (lead all the features close to zero), we replace the ImageNet pre-trained model in RC by exponential moving average (EMA) model. Comparing the fifth row and the sixth row in Tab.~\ref{table:ablation-loss}, EMA model only gains 1.25\% improvement while RC improves the SC model by 2.75\%. The results verify the significance of the retrospective knowledge in RC.

\noindent\textbf{Ablation Studies on Image Classification.}
We further conduct ablation studies on the PACS benchmark to verify the effectiveness of each component in \ours.
Different phenomena from the segmentation benchmark are observed. \textit{First}, adopting SHM can only slightly improve the model performance since the styles of a single classification dataset are limited. \textit{Second}, the performance yields an significant improvement of 4.6\% in average accuracy when style consistency is applied. \textit{Third}, leveraging retrospective consistency can gain an improvement of 1.2\% and applying it on top of the style consistency can further improve the performance by 1.6\% in average accuracy.

\subsection{Further Evaluation}
In this section, we compare different style variation methods under the proposed dual consistency constraints and analyze two important hyper-parameter, \ie, the location of SHM and the basis style selection interval, in our framework. Experiments in this section are based on the synthetic-to-real single-source DG in semantic segmentation. ResNet-50~\citep{he2016deep} is leveraged as the backbone.

\begin{figure*}[ht]
    \centering
     \begin{subfigure}[ht]{0.9\linewidth}
         \centering
         \includegraphics[width=\textwidth]{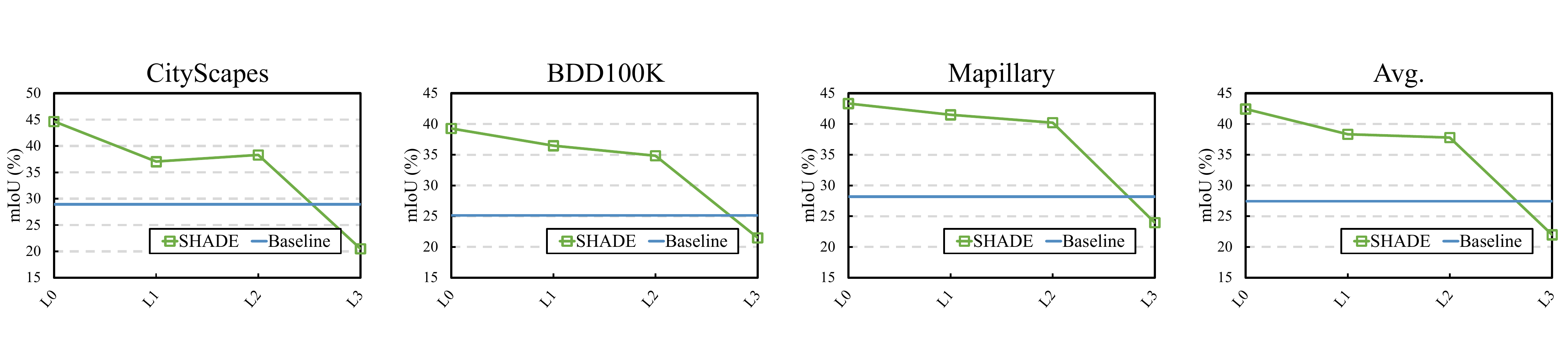}
         \vspace{-.3in}
         \caption{Location of SHM.}
         \label{fig:param-loc}
     \end{subfigure}
     \\

     \begin{subfigure}[ht]{0.9\linewidth}
         \centering
         \includegraphics[width=\textwidth]{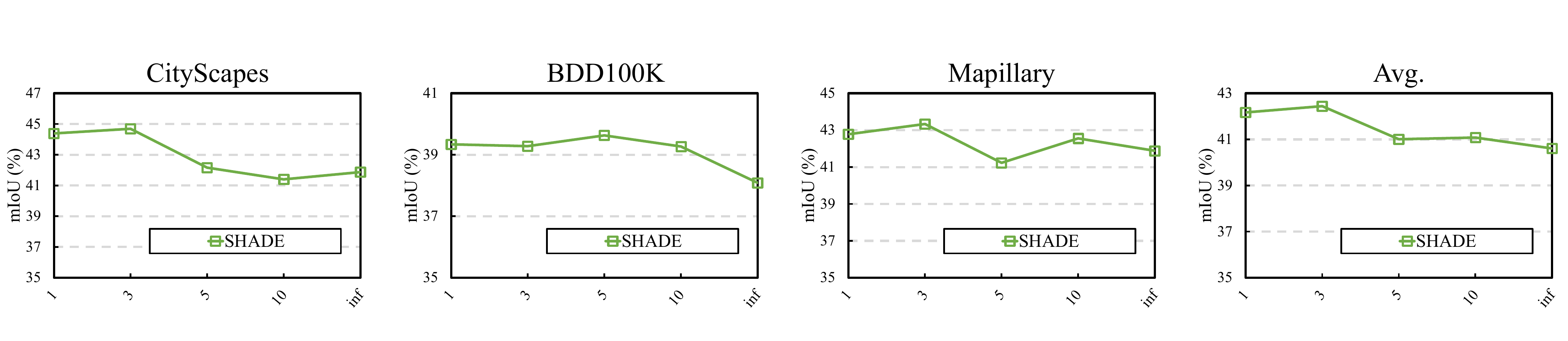}
         \vspace{-.3in}
         \caption{Basis style selection interval.}
         \label{fig:param-interval}
     \end{subfigure}
     \vspace{-.1in}
    \caption{Parameter analysis.}
    \label{fig:param}
\end{figure*}

\begin{figure*}[ht]
    \centering
    \includegraphics[width=.95\linewidth]{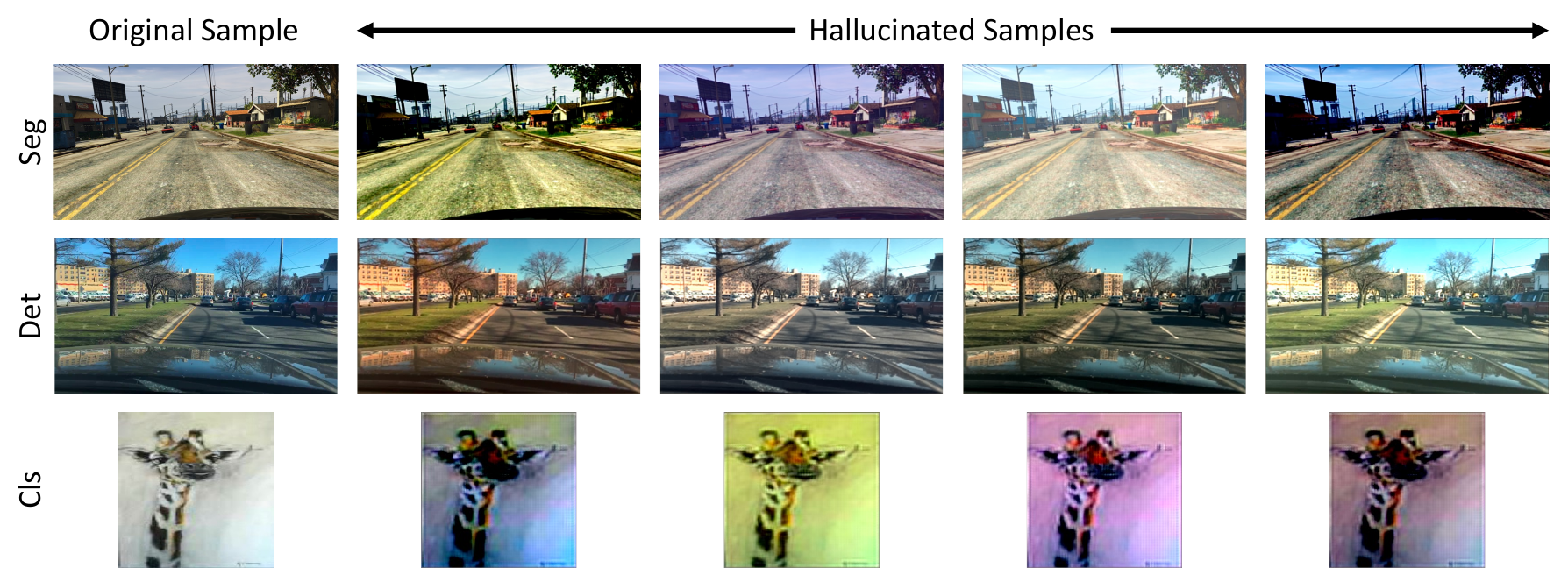}
    \caption{Visualization of style-diversified samples.}
    \label{fig:visual-style}
\end{figure*}

\begin{figure*}[ht]
    \centering
    \includegraphics[width=.95\linewidth]{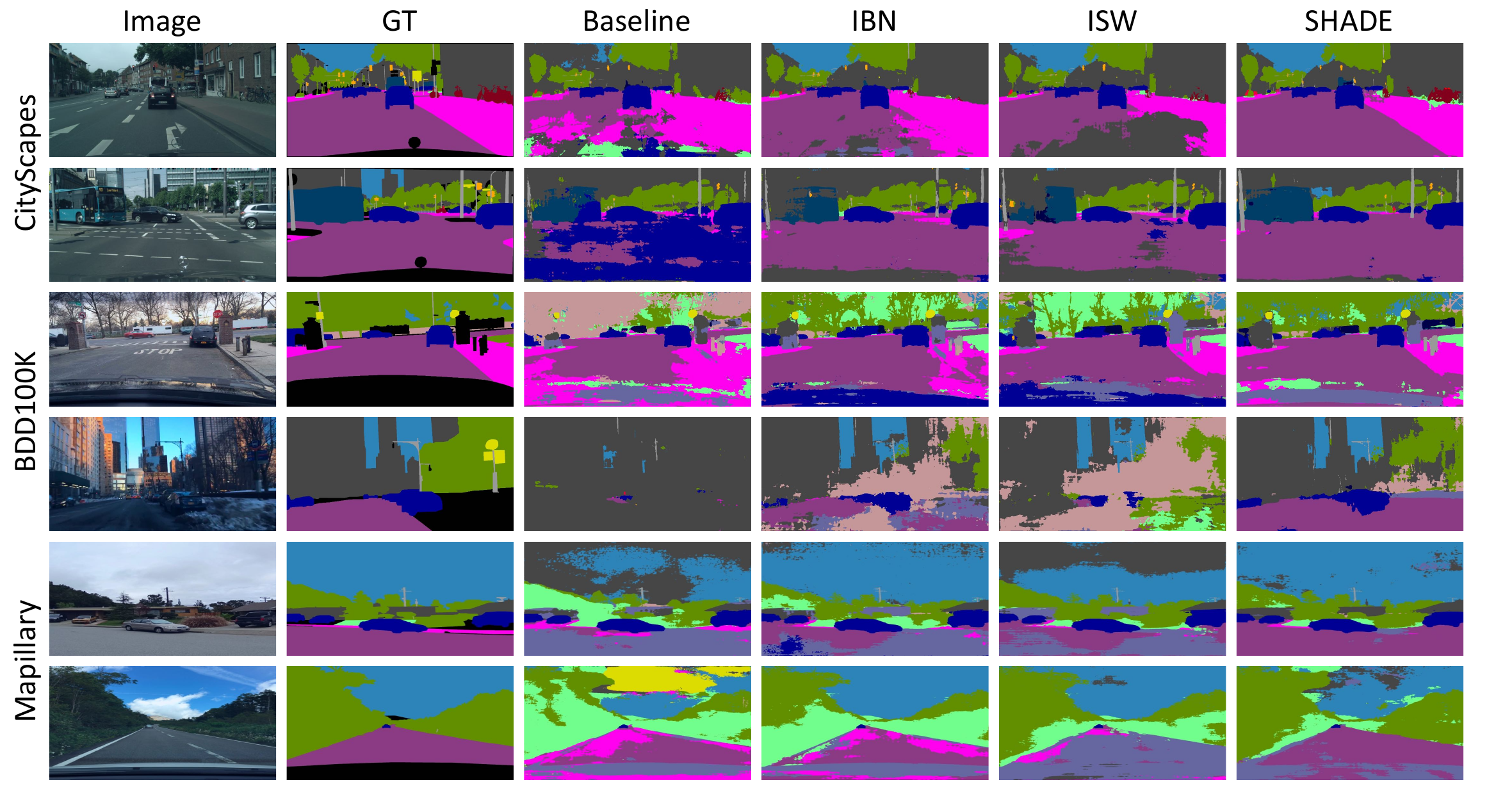}
    \vspace{-.1in}
    \caption{Qualitative comparison of segmentation results.}
    \label{fig:seg_results}
\end{figure*}

\begin{figure}[ht]
    \centering
    \includegraphics[width=.95\linewidth]{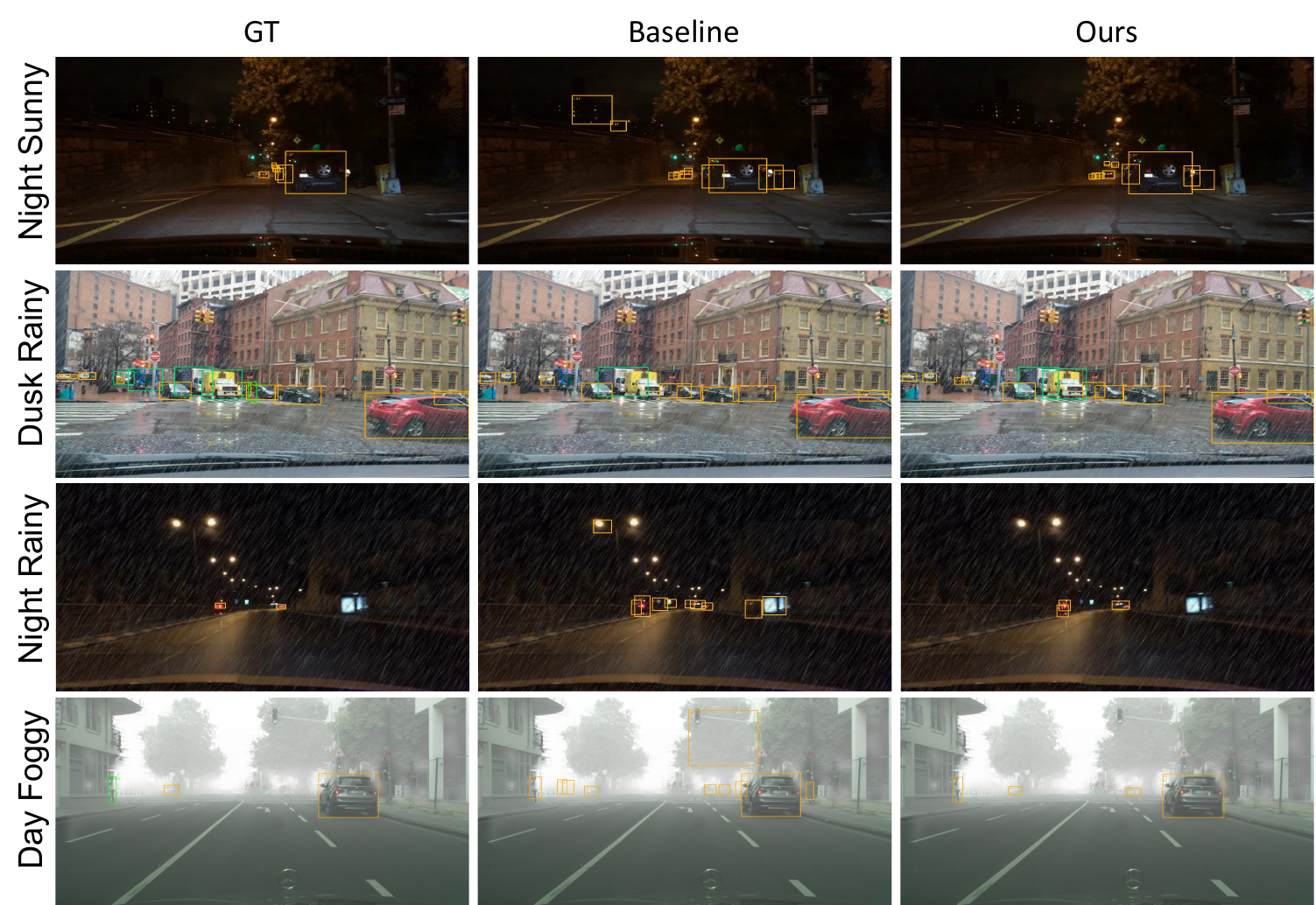}
    \caption{Qualitative comparison of detection results.}
    \label{fig:det_results}
\end{figure}

\noindent\textbf{Comparison of different style variation methods.}
We compare SHM with random style, MixStyle~\citep{zhou2021mixstyle}, CrossNorm~\citep{crossnorm} and style hallucination with Kmeans basis in Tab.~\ref{table:style-variation}. Random style utilizes the sampled styles from the standard normal distribution to form new samples.
MixStyle~\citep{zhou2021mixstyle} generates new styles by mixing the original style with the randomly shuffled style within a mini-batch, and CrossNorm~\citep{crossnorm} swaps the original style with another style within the shuffled mini-batch. 
SHM and Kmeans basis both use the linear combination of basis style to generate new styles, but the basis styles of SHM are selected by FPS~\citep{qi2017pointnet++} while those of Kmeans basis are obtained by Kmeans clustering centers~\citep{macqueen1967some_kmeans}. We can make four observations from Tab.~\ref{table:style-variation}.
\textit{First}, we cannot make full use of dual consistency to achieve significant performance with the unrealistic random styles since standard normal distribution cannot represent the source nor the target domains.
\textit{Second}, despite the use of realistic source styles, random utilization of MixStyle and CrossNorm leads to the generation of more samples from the dominant styles that may be different from the target styles. When using MixStyle and CrossNorm, the model achieves an average mIoU of 41.10\% and 40.72\%, respectively. 
\textit{Third}, as shown in Fig.~\ref{fig:visual-dist}(b), Kmeans basis suffers from the similar but more severe dominant style issue in style generation. As a result,
rare styles are almost discarded and thus the model achieves poorer performance than the above two.
\textit{Fourth}, SHM selects basis styles via FPS, and thus the selection can cover the source distribution to  
a large extent, especially those rare styles. With such basis styles, SHM generates styles from all the source distributions, and some generated styles are even close to the target domains (Fig.~\ref{fig:visual-dist}(c)). Consequently, combining SHM with dual consistency learning, \ours can reap the benefit of the source data and outperform other methods on all three target datasets.

\noindent\textbf{Location of SHM.}
We investigate the impact of inserting SHM in different locations in Fig.~\ref{fig:param-loc}. ``L0'' denotes inserting SHM after the first Conv-BN-ReLU layer (layer0) and ``L1'' to ``L3'' denote inserting SHM after the corresponding (1-3) ResNet layer. As shown in Fig.~\ref{fig:param-loc}, ``L0'' achieves the best result while the performance of ``L1'' and ``L2'' drops a little. However, the model suffers from drastic performance degradation when inserting SHM after layer3, and the performance is even worse than the baseline. The reasons are two-fold. \textit{First}, the channel-wise mean and standard deviation represent more style information in the shallow layers of deep neural networks while they contain more semantic information in deep layers~\citep{adain,dumoulin2016learned}. \textit{Second}, the residual connections in ResNet will lead the ResNet activations of deep layers to have large peaks and small entropy, which makes the style features biased to a few dominant patterns instead of the global style~\citep{wang2021rethinking}.
Based on the above observations, we insert SHM after the shallow layer of the backbone, \ie, layer0 of the ResNet~\citep{he2016deep} and block1 of MiT-B5~\citep{xie2021segformer}.

\noindent\textbf{Basis Style Selection Interval.}
The distribution of source styles is varied along with the model training. 
To better represent the style space, we re-select the basis styles with the interval of $k$ epochs. 
The abscissa of Fig.~\ref{fig:param-interval} denotes the selection interval $k$ and ``inf'' denotes only selecting the basis style once in the beginning of training.
As shown in Fig.~\ref{fig:param-interval}, the model achieves consistent and good performance with frequent re-selection ($k<=3$) while the performance degrades with the increase of selection interval, and the average mIoU is lower than 41\% when only selecting once. Taking both the performance and computational cost into consideration, we set $k=3$ in DG-Seg with ResNet backbone.
\textbf{Note that} the batch size and training data are different across the models and tasks, and thus we follow the frequent re-selection observation to set the frequency of DG-Seg with MiT-B5 backbone and DG-Det to 4k iterations and 1 epoch, respectively.

\subsection{Visualization}

\noindent\textbf{Style Visualization.} To better understand our SHM, we visualize the style-diversified samples with an auto-encoder. We take the fixed pre-trained blocks before SHM as the encoder, and add an additional decoder. The decoder is trained by L1 loss with the source data. After training, we use SHM to generate multiple styles and replace the style of the original sample. Then we use decoder to generate the visualization results. 
Examples of the three tasks are shown in Fig.~\ref{fig:visual-style}. 
SHM replaces the original style features with the combination of basis styles to obtain new samples of different styles, \eg, weather change and time change.

\noindent\textbf{Qualitative results.} To demonstrate the effectiveness of \ours, we compare the qualitative results of semantic segmentation and object detection.
We compare the segmentation results among baseline, IBN-Net~\citep{ibn}, ISW~\citep{robustnet} and \ours on CityScapes~\citep{cityscapes}, BDD100K~\citep{bdd} and Mapillary~\citep{mapillary} in Fig.~\ref{fig:seg_results}. 
We obtain two observations from Fig.~\ref{fig:seg_results}.
\textit{First}, \ours consistently outperforms other methods under different target conditions (\eg, sunny, cloudy and overcast).
\textit{Second}, \ours can well deal with both background classes (\eg, road) and foreground classes (\eg, bus and bicycle).
In addition, we compare \ours with the baseline model on object detection benchmark in Fig.~\ref{fig:det_results} and \ours consistently outperforms the baseline under different environmental conditions.
The above observations demonstrate that \ours is robust to style variation and has strong ability in addressing unseen images.

\section{Conclusion}
\label{sec:conclusion}

In this paper, we present a novel framework~(\ours) for visual domain generalization. To address the distribution shift between the source and unseen target domains, \ours leverages two consistency constraints to learn the domain-invariant representation by seeking consistent representation across styles and the guidance of retrospective knowledge. 
In addition, the style hallucination module (SHM) is equipped into our framework, which can effectively catalyze the dual consistency learning by generating diverse and realistic source samples. 
The proposed \ours is effective and versatile, which can be applied to image classification, semantic segmentation and object detection tasks with both ConvNets and Transformer backbone, and can achieve state-of-the-art performance on different benchmarks and under different settings.

\section{Acknowledgements}
This research/project is supported by the National Research Foundation Singapore and DSO National Laboratories under the AI Singapore Programme (AISG Award No: AISG2-RP-2020-016), the Tier 2 grant MOE-T2EP20120-0011 from the Singapore Ministry of Education, the MUR PNRR project FAIR (PE00000013) funded by the NextGenerationEU, and the EU project Ai4Trust (No. 101070190).

\bibliographystyle{spbasic}      
\bibliography{ref}   

\end{document}